\documentclass[runningheads]{llncs}
\usepackage[T1]{fontenc}
\usepackage{graphicx}

\usepackage{cuted}
\usepackage{amsmath}
\usepackage{hyperref}
\usepackage{pifont}
\usepackage{marvosym}
\usepackage{amssymb}
\usepackage{subcaption}
\usepackage{booktabs}
\usepackage{multirow}
\usepackage{graphicx}
\usepackage{xcolor}
\usepackage{placeins}
\usepackage[table]{xcolor}
\usepackage[normalem]{ulem}

\usepackage{wrapfig}

\begin{document}

\title{T-araVLN: Translator for Agricultural Robotic Agents on Vision-and-Language Navigation}

\titlerunning{T-araVLN · Preprint}




\author{
Xiaobei Zhao \and
Xingqi Lyu \and
Xin Chen\textsuperscript{\Letter} \and
Xiang Li\textsuperscript{\Letter}
}

\authorrunning{Xiaobei Zhao et al.}

\institute{China Agricultural University\\
\email{\{xiaobeizhao2002,lxq99725\}@163.com, \{chxin,cqlixiang\}@cau.edu.cn}}

\maketitle

\begin{abstract}
Agricultural robotic agents have been becoming useful helpers in a wide range of agricultural tasks. However, they still heavily rely on manual operations or fixed railways for movement. To address this limitation, the AgriVLN method and the A2A benchmark pioneeringly extend Vision-and-Language Navigation (VLN) to the agricultural domain, enabling agents to navigate to the target positions following the natural language instructions. We observe that AgriVLN can effectively understands the simple instructions, but often misunderstands the complex ones. To bridge this gap, we propose the T-araVLN method, in which we build the instruction translator module to translate noisy and mistaken instructions into refined and precise representations. When evaluated on A2A, our T-araVLN successfully improves Success Rate (SR) from 0.47 to 0.63 and reduces Navigation Error (NE) from 2.91m to 2.28m, demonstrating the state-of-the-art performance in the agricultural VLN domain. 
Code: \href{https://github.com/AlexTraveling/T-araVLN}{https://github.com/AlexTraveling/T-araVLN}.

\keywords{Vision-and-Language Navigation \and Agricultural Robotic Agent \and Large Language Model.}
\end{abstract}
\section{Introduction}
\label{sec:introduction}

\begin{figure}[t]
\centering
\includegraphics[width=0.75\linewidth]{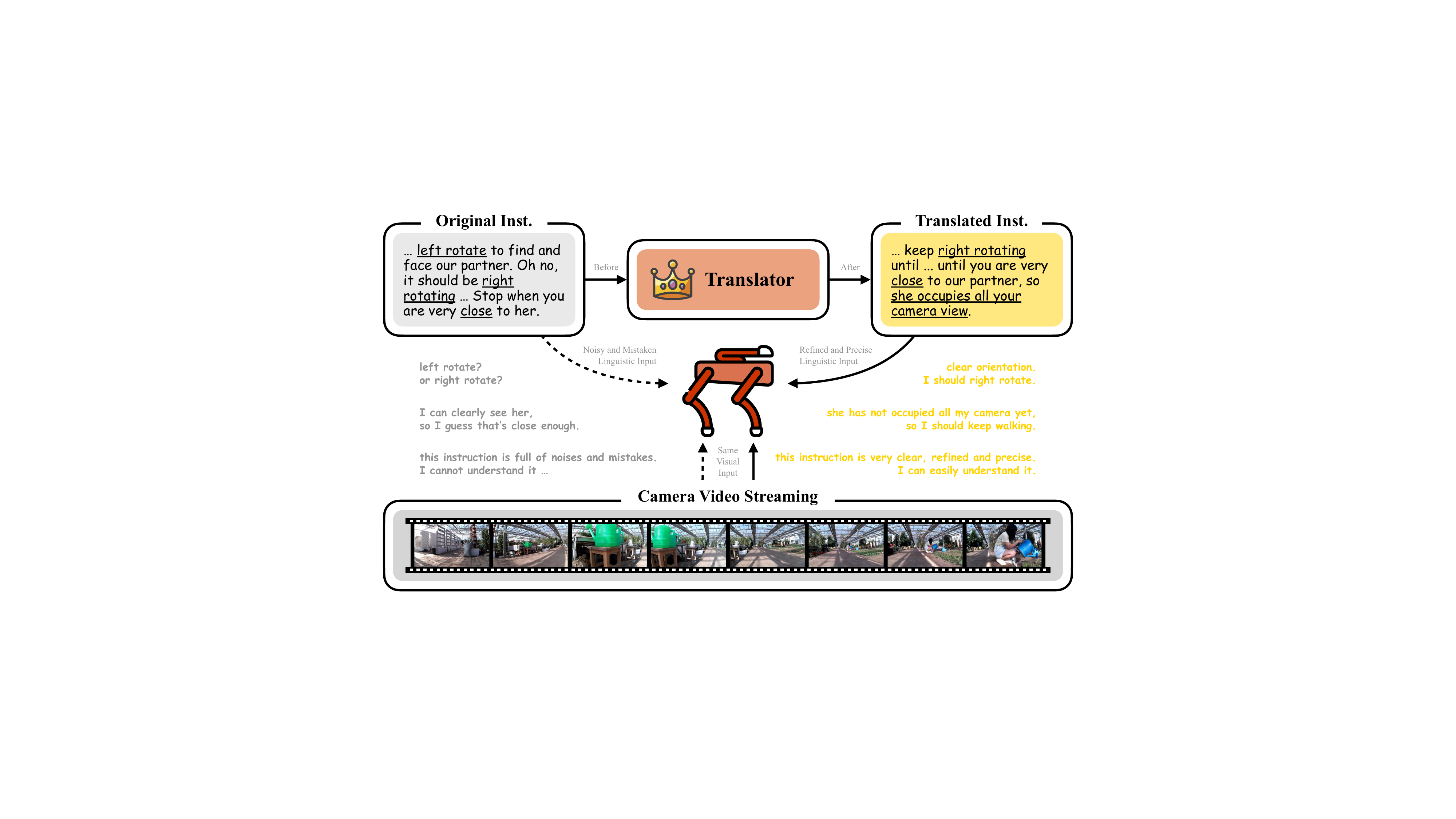}
\caption{
T-araVLN (right) v.s. the baseline (left) on a simple example: Noisy and mistaken instructions such as \textit{“close”} are translated to 
refined and precise representations 
such as \textit{“she occupies all your camera view”}, leading to an easier alignment between linguistic and visual inputs 
for the decision-maker.
}
\label{fig:teaser}
\end{figure}

\par Agricultural robotic agents have been becoming useful helpers in a wide range of agricultural tasks, such as 
laser weeding \cite{COMPAG:LaserWeeding}, growth monitoring \cite{RAL:GrowthMonitoring} and cross-pollination \cite{Cell:Pollination}. 
However, most of them still heavily rely on manual operations or fixed railways for movement, resulting in limited mobility and poor adaptability on diversified scenarios.

\par In contrast, Vision-and-Language Navigation (VLN) enables agents to follow the natural language instructions to navigate to the target positions \cite{CVPR:R2R,ECCV:VLN-CE}, having demonstrated strong performance across various domains \cite{TMLR}, such as R2R \cite{CVPR:R2R} for indoor rooms, TOUCHDOWN \cite{CVPR:TOUCHDOWN} for urban streets, and AerialVLN \cite{ICCV:AerialVLN} for aerial spaces. Motivated by prior Vision-Language Model-based methods \cite{AAAI:NavGPT,ECCV:NavGPT-2,CVPR:Long}, Zhao et al. \cite{arXiv:AgriVLN} proposes the AgriVLN method and the A2A benchmark, pioneeringly extending VLN to the agricultural domain. To better align with the practical speaking tone of agricultural workers, A2A's instructions are deliberately designed with many noises and mistakes, which inevitably brings significant challenges for the alignment between linguistic and visual inputs \cite{ICLR:GSA-VLN}. Consequently, we observe that AgriVLN effectively understands the simple instructions, but often misunderstands the complex ones, as illustrated in the left part of Figure \ref{fig:teaser}. This interesting phenomenon brings us a question: \textit{\textbf{Can we polish an instruction to make it easier to understand?}}

\par The most relevant study, as far as we know, is VLN-Trans \cite{ACL:VLN-Trans}, which trains a translator to convert original instructions into easy-to-follow sub-instruction representations, 
performing well on the indoor benchmarks \cite{CVPR:R2R,ACL:R4R,NeurIPS:HAMT}.
However, the instruction issue types in A2A are much more complicated than in the indoor benchmarks, which makes training-based methods like VLN-Trans \cite{ACL:VLN-Trans} and MTST \cite{ACL:MTST} struggling with the circumstances out of the training data, leaving a large gap on generalization capability.

\par To bridge this gap, we propose the method of \textbf{T}ranslator for \textbf{A}gricultural \textbf{R}obotic \textbf{A}gents on \textbf{V}ision-and-\textbf{L}anguage \textbf{N}avigation (\textbf{T-araVLN}), as illustrated in Figure \ref{fig:teaser}. First, we prompt a Large Language Model (LLM) in a few-shot manner to build the module of instruction translator, to translate noisy and mistaken instructions into refined and precise representations. Second, we integrate it into the base model to establish our T-araVLN method. Third, we evaluate T-araVLN on the A2A benchmark, which effectively improves Success Rate (SR) from 0.47 to 0.63 and reduces Navigation Error (NE) from 2.91m to 2.28m, demonstrating the state-of-the-art performance in the agricultural VLN domain. 
\par We share a simple example to better explain the difference between T-araVLN and the baseline, as illustrated in Figure \ref{fig:teaser}. In the original instruction, the user gives several unspecific descriptions, such as \textit{“stop when you are very close to her”}, in which the decision-maker cannot understand how \textit{“close”} the distance is the best, so wrongly regards \textit{“I can clearly see her”} as \textit{“close enough”}, resulting in the over early \texttt{[STOP]}. When the translator is integrated, it translates the noisy and mistaken contents, such as \textit{“close”}, to be more refined and precise, such as \textit{“she occupies all your camera view”}, helping the decision-maker easily recognize that \textit{“she has not occupied all my camera yet”} so properly \textit{“keep walking} \texttt{[FORWARD]}\textit{”}.

\par In summary, our main contributions are as follows:
\begin{itemize}
\item \textbf{Instruction Translator}, a LLM-based linguistic preprocessing module, translates noisy and mistaken instructions into refined and precise representations. 
\item \textbf{T-araVLN}, a VLN method integrating the instruction translator, navigates agricultural robotic agents to the target positions following the natural language instructions. 
\item We implement the ablation study and qualitative experiment showing the effectiveness of the instruction translator, and the comparison experiment showing the state-of-the-art performance of T-araVLN.
\end{itemize}
\par Codes will be available after acceptance.

\section{Related Works}
\label{sec:related_works}

\subsection{The A2A Benchmark}
\par Agriculture-to-Agriculture (A2A) \cite{arXiv:AgriVLN} is currently the only one VLN benchmark specially designed for agricultural robots, consisting of 1,560 episodes across 6 types of scene: farm, greenhouse, forest, mountain, garden and village, in which all the instructions belong to the step-by-step format and the action space belongs to the continuous environment.

\subsection{The AgriVLN Method}
Vision-and-Language Navigation for Agricultural Robots (AgriVLN) \cite{arXiv:AgriVLN} is the first agricultural VLN method, which uses NavGPT \cite{AAAI:NavGPT} as the backbone and integrate the LLM-based Subtask List (STL) module, enabling an agricultural robotic agent navigating to the target positions following the natural language instructions.


\subsection{Instruction Translation in Vision-and-Language Navigation}
Recent studies have explored enhancing the VLN performance through diversified patterns, such as historical memory representation \cite{ICRA:VLN-KHVR,ACL:CityNavAgent,AAAI:HETT}, dynamic topological mapping \cite{ICCV:GridMM,ACL:MapGPT,IROS:Map}, and multi-experts collaboration \cite{ICRA:DiscussBeforeMoving,KBS,ICCV:SAME}, effectively achieving varying degrees of improvement. Nevertheless, we observe that the instruction quality attracts less attention from researchers. The most relevant study discussing instruction translation mechanism, as far as we know, is VLN-Trans \cite{ACL:VLN-Trans}. It trains a LSTM-based \cite{LSTM} translator taking the original instruction and the current viewpoints as inputs to generate new sub-instruction representations, in which one positive, one anchor and three negative labels are annotated to build the triplet loss function.
It performs well on the three indoor benchmarks \cite{CVPR:R2R,ACL:R4R,NeurIPS:HAMT}, but we observe two limitations: 1) The annotation is laborious but necessary, resulting in a difficult transfer to other domains, such as aerial and agricultural scenes; 2) The architecture is based on training, resulting in a poor generalization capability on the scenarios out of the training data. 
Hence, we suggest that a novel training-free instruction translator is needful for VLN in agriculture.

\section{Methodology}
\label{sec:methodology}
\begin{figure*}[t]
\centering
\includegraphics[width=1.0\linewidth]{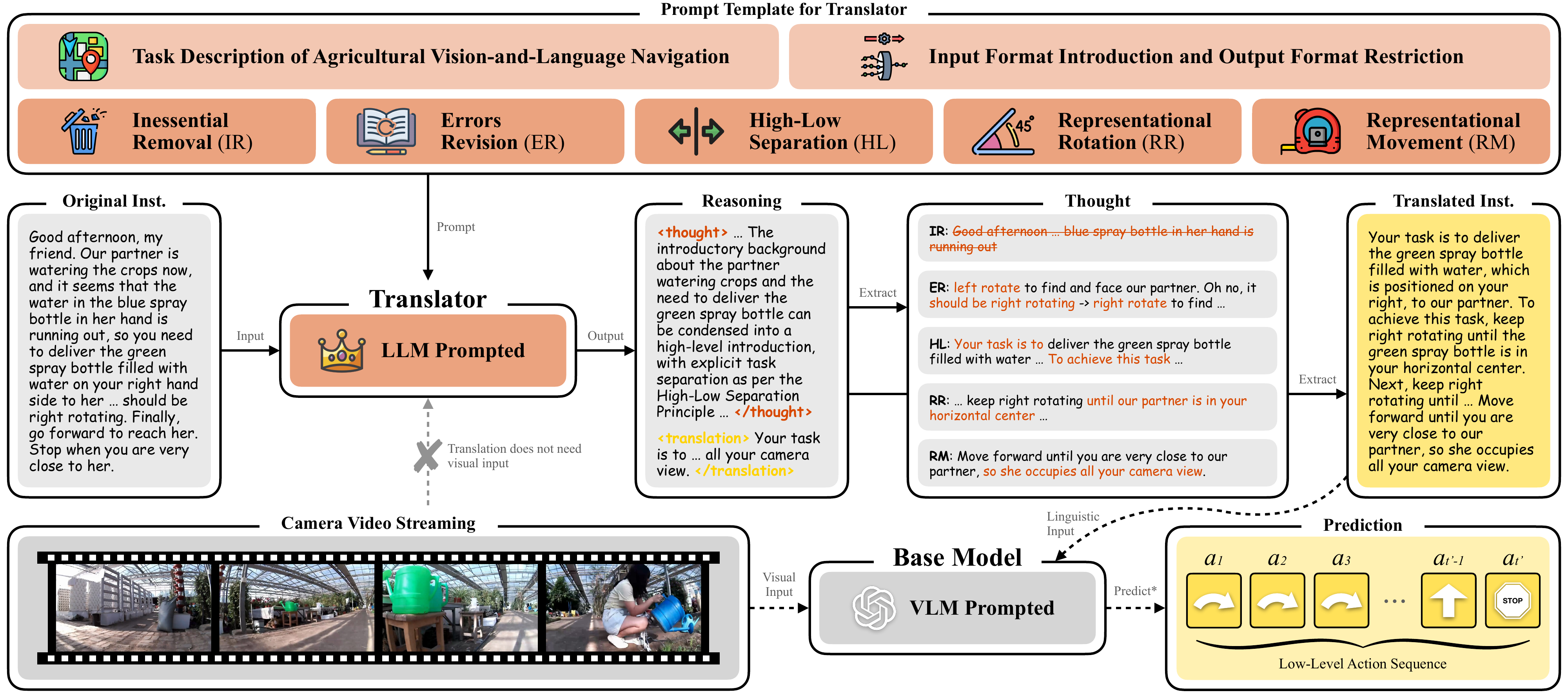}
\caption{
T-araVLN methodology illustration: Prompted by the five translation principles with the task description and the format restriction, T-araVLN translates the original instruction to be more refined and precise, then understands both linguistic and visual inputs to predict the low-level action sequence, navigating the agent to move from the starting point to the target position. 
(*The base model's elaborate prediction process is detailed in AgriVLN \cite{arXiv:AgriVLN})}
\label{fig:method}
\end{figure*}

In this section, we present the T-araVLN method, as illustrated in Figure \ref{fig:method}. First, we introduce the task definition of agricultural VLN in Sec. \ref{sec:task_definition}. Second, we present the instruction translator module in Sec. \ref{sec:instruction_translator}. Third, we integrate it into the base model to build T-araVLN in Sec. \ref{sec:base_model}.

\subsection{Task Definition}
\label{sec:task_definition}
The task of Agricultural Vision-and-Language Navigation \cite{arXiv:AgriVLN} is defined as follows: In each episode, the model is given an instruction in natural language, denoted as $W = \langle w_1, w_2, \dots, w_L \rangle$, where $L$ is the number of words. At each time step $t$, the model is given the front-facing RGB image $I_t \in \mathbb{R}^{3 \times \mathbb{H} \times \mathbb{W}}$, where $\mathbb{H}$ and $\mathbb{W}$ are set to 360 and 640 as the default values, respectively. The purpose is understanding both $W$ and $I_t$ to predict the most appropriate low-level action $\hat{a_t}$ from the action space $\{ \texttt{[FORWARD]}$, $\texttt{[LEFT ROTATE]}$, $\texttt{[RIGHT ROTATE]}$, $\texttt{[STOP]} \}$, thereby navigating the agricultural robotic agent to move from the starting point to the target position.

\subsection{Instruction Translator}
\label{sec:instruction_translator}
\par We prompt a Large Language Model (LLM) in a few-shot manner to build the instruction translator module, denoted as $\mathcal{T}\,(\,\cdot\,)$. For an original instruction $W$ in a casual spoken-language style containing complete semantics, $\mathcal{T}\,(\,\cdot\,)$ translates it to a formal written-language style retaining only essential semantics, defined as:
\begin{equation}
W', \rho = \mathcal{T}\,\big(\boldsymbol{\tau}^{\text{sys}}_{\mathcal{T}}, \boldsymbol{\tau}^{\text{usr}}_{\mathcal{T}} (W)\big)
\end{equation}
where $\boldsymbol{\tau}^{\text{sys}}_{\mathcal{T}}$ and $\boldsymbol{\tau}^{\text{usr}}_{\mathcal{T}}$ are the system and user prompt templates for $\mathcal{T}\,(\,\cdot\,)$, respectively. $W' = \langle w'_1, w'_2, \dots, w'_{L'} \rangle$ is the instruction after translation, where $L'$ is the number of words. $\rho$ is the reasoning thought, providing an explicit interpretation. 
\par To support the translation, we design a total of five translation principles, denoted as $\{ P_i \}_{i=1}^5
$, in which each principle $P_i$ is represented in an one-shot manner, 
consisting of a concise description paired with an illustrative example.
\begin{enumerate}
  \renewcommand{\labelenumi}{\textbf{P\arabic{enumi}.}}

  \item \textbf{Inessential Removal (IR).}
  Delete all the inessential words which have no direct connection to navigation, such as “good morning”.
  
  \item \textbf{Errors Revision (ER).}
  Revise all the speaking errors, such as “left rotate to face it... no, right rotate” should be revised to “right rotate to face it”.

  \item \textbf{High-Low Separation (HL).}
  Distinguish between the high-level introduction (if present) and the low-level orders, then clearly separate them using an explicit delimiter phrase, such as “you need to navigate to the water station to get some water so that we can water plants. right rotate to...” can be revised to “your task is to navigate to the water station. To achieve this task, you need to right rotate to...”.

  \item \textbf{Representational Rotation (RR).}
  Polish the end condition for rotation (if present) from abstract description to representational description, such as “left rotate to face the green sprayer” can be revised to “keep left rotating to face the green sprayer until it is in your horizontal center”.

  \item \textbf{Representational Movement (RM).}
  Polish the end condition for moving forward (if present) from abstract description to representational description, such as “stop when you reach the blue tractor” can be revised to “stop when you are very close to the blue tractor, at that time it should occupy all your camera view”.
\end{enumerate}
We package all the translation principles into the system prompt template, defined as:
\begin{equation}
\boldsymbol{\tau}^{\text{sys}}_{\mathcal{T}}
\;\triangleq\;
\big(
\tau_{\text{base}}
\;\oplus\;
\langle P_i \rangle_{i=1}^5
\;\oplus\;
\tau_{\text{fmt}}
\big)
\end{equation}
where $\tau_{\text{base}}$ is the basic description. $\tau_{\text{fmt}}$ is the format restriction on input and output. $\oplus$ is the semantic concatenation.
After translation, $W'$ is easier than $W$ for the downstream decision-maker to understand.

\subsection{Base Model}
\label{sec:base_model}
\par We follow the architecture of AgriVLN \cite{arXiv:AgriVLN} as our base model, in which we integrate the instruction translator module as a linguistic preprocessing layer, then we prompt a Vision-Language Model (VLM) in an end-to-end paradigm to build the decision-maker, denoted as $\mathcal{D}\,(\,\cdot\,)$. 
At each time step $t$, $\mathcal{D}\,(\,\cdot\,)$ takes the translated instruction $W'$ as the linguistic input and the camera image $I_t$ as the visual input, to predict the most appropriate low-level action $\hat{a_t}$, defined as: 
\begin{equation}
\hat{a_{t}}, \rho_t = \mathcal{D}\,\big(\boldsymbol{\tau}^{\text{sys}}_{\mathcal{D}}, \boldsymbol{\tau}^{\text{usr}}_{\mathcal{D}} (W', I_{t})\big)
\end{equation}
where $\boldsymbol{\tau}^{\text{sys}}_{\mathcal{D}}$ and $\boldsymbol{\tau}^{\text{usr}}_{\mathcal{D}}$ are the system and user prompt templates for $\mathcal{D}\,(\,\cdot\,)$, respectively. $\rho_t$ is the reasoning thought, providing an explicit interpretation. 
In an episode, $\mathcal{D}\,(\,\cdot\,)$ ends when one of the following conditions happens: 
\begin{quote}
1) $\hat{a_{t'}}$ = $\texttt{[STOP]}$; \\
2) The predicted action sequence $\langle \hat{a_{t'-\delta}}, \hat{a_{t'-\delta+1}}, \dots, \hat{a_{t'}} \rangle$ is deviated to the ground-truth action sequence $\langle a_{t'-\delta}, a_{t'-\delta+1}, \dots, a_{t'} \rangle$; \\
3) $t'$ reaches the max limitation of time step.
\end{quote}
where $t'$ is the ending time step. $\delta$ is the deviation time threshold, which is set to 4s following A2A \cite{arXiv:AgriVLN}'s setting.


\section{Experiments}
\label{sec:experiments}

\subsection{Experimental Settings}
We implement all the experiments on the A2A \cite{arXiv:AgriVLN} benchmark. 
In our proposed T-araVLN method, we follow GPT-4.1 mini \cite{gpt} as the default Vision-Language Model (VLM) for decision-making, and select GPT-4.1 \cite{gpt} as the default Large Language Model (LLM) for instruction translation. 
We access all the VLMs and LLMs through their official APIs. 
All the experiments run on a single NVIDIA L20 GPU \cite{NVIDIA} with 48G video memory.



\begin{table*}[t]
\caption{Comparison experiment results between T-araVLN and the state-of-the-art methods on the low-complexity portion (subtask $=$ 2), the high-complexity portion (subtask $\geq$ 3), and the whole of the A2A \cite{arXiv:AgriVLN} benchmark.}
\label{tab:comparison_experiment}
\vspace{2.0ex}

\centering
\resizebox{\linewidth}{!}{
\renewcommand{\arraystretch}{1.1}
\begin{tabular}{rll ccc|ccc|ccc}
\toprule
\multirow{2}{*}{\textbf{\#}} & \multirow{2}{*}{\textbf{Method}} & \multirow{2}{*}{\textbf{LLM for T.}} & 
\multicolumn{3}{c}{\textbf{A2A ( low )}} & 
\multicolumn{3}{c}{\textbf{A2A ( high )}} & 
\multicolumn{3}{c}{\textbf{A2A}} \\
\cmidrule(lr){4-6} \cmidrule(lr){7-9} \cmidrule(lr){10-12}
& & & \textbf{SR}$\uparrow$ & \textbf{NE}$\downarrow$ & \textbf{BERTScr.} & 
\textbf{SR}$\uparrow$ & \textbf{NE}$\downarrow$ & \textbf{BERTScr.} & 
\textbf{SR}$\uparrow$ & \textbf{NE}$\downarrow$ & \textbf{BERTScr.} \\
\midrule
1 & Random   & -                       & 0.13 & 7.30          & - & 0.04 & 6.74 & - & 0.09 & 7.03 & - \\
2 & Fixed    & -                       & 0.00 & 0.00          & - & 0.06 & 6.32 & - & 0.03 & 3.06 & - \\
\midrule \rowcolor{gray!15} \multicolumn{12}{c}{\textit{State-of-the-Art}} \\
3 & NavGPT    & - & 0.51             & \textbf{0.60}    & -      & 0.14             & 5.01             & -      & 0.33             & 2.76             & - \\
4 & SIA-VLN   & - & 0.52             & 1.46             & -      & \underline{0.08} & \underline{5.12} & -      & \underline{0.31} & \underline{3.24} & - \\
5 & DILLM-VLN & - & \underline{0.41} & 1.36             & -      & 0.32             & 3.90             & -      & 0.36             & 2.60             & - \\
6 & AgriVLN   & - & 0.58             & \underline{2.32} & -      & 0.35             & 3.54             & -      & 0.47             & 2.91             & - \\
\midrule \rowcolor{gray!15} \multicolumn{12}{c}{\textit{Ours}} \\
7 & T-araVLN$_{c}$ & Claude-3.7 & 0.66             & 2.12             & 0.9096 & \textbf{0.49}    & 3.88             & 0.9112 & 0.58             & 2.97             & 0.9105 \\
8 & T-araVLN$_{d}$ & DeepSeek-R1       & 0.70             & 1.68             & 0.8977 & 0.47             & 3.49             & 0.8982 & 0.59             & 2.55             & 0.8980 \\
9 & T-araVLN$_{g}$ & GPT-4.1           & \textbf{0.80}    & 1.41             & 0.9028 & 0.46             & \textbf{3.22}    & 0.9039 & \textbf{0.63}    & \textbf{2.28}    & 0.9035 \\
\midrule
10 & Human    & -                       & 0.93 & 0.32 & - & 0.80 & 0.82 & - & 0.87 & 0.57 & - \\
\bottomrule
\end{tabular}
}

\footnotesize 
\vspace{2.0ex}
$\uparrow$ and $\downarrow$ indicate that higher and lower values correspond to better performance, respectively. 
T. and BERTScr. represent the instruction translator and BERTScore, respectively. 
\textbf{Bold} and \underline{underline} mark the best and worst scores, respectively.

\end{table*}

\subsection{Evaluation Metrics}
We follow the two standard evaluation metrics for VLN: \textbf{S}uccess \textbf{R}ate (\textbf{SR}) and \textbf{N}avigation \textbf{E}rror (\textbf{NE}) \cite{CVPR:R2R}. NE measures the path length between the stopping position and the target position. SR measures the rate successfully reaching the target position within a 2-meter NE. Besides, we introduce \textbf{BERTScore$_{F1}$} \cite{ICLR:BERTScore} to calculate the semantic similarity between the original instruction and the translated one, in which a lower score indicates a greater degree of modification, leading to a more aggressive translation preference; conversely, a higher score indicates a smaller degree of modification, leading to a more conservative translation preference. Please note that BERTScore$_{F1}$ reflects only the translation preference, but not the translation quality. 

\subsection{Comparison Experiment}
\par We compare T-araVLN with four existing state-of-the-art models: NavGPT \cite{AAAI:NavGPT}, SIA-VLN \cite{EMNLP:SIA-VLN}, DILLM-VLN \cite{RAL:DILLM-VLN} and AgriVLN \cite{arXiv:AgriVLN}. Besides, the methods of Random, Fixed and Human are reproduced as the lower and upper bounds, respectively. 
The results are shown in Table \ref{tab:comparison_experiment}. On the whole of A2A, our T-araVLN with GPT-4.1-based translator (\#9) achieves SR of 0.63 and NE of 2.28m, outperforming all the four models (\#3, \#4, \#5 and \#6) on both SR and NE. On the low-complexity portion, our T-araVLN even substantially improves SR from 0.58 to 0.80, which nearly approaches the performance of human. In summary, the comparison experiment demonstrates the significant effectiveness of the instruction translator module, which leads to T-araVLN's state-of-the-art performance in the agricultural VLN domain. 


\subsection{Ablation Study}
\subsubsection{Large Language Model}
We test T-araVLN with two additional LLMs of comparable size and performance, as shown in Table \ref{tab:comparison_experiment}. When the instruction translator module is driven by Claude-3.7-Sonnet \cite{claude} (\#7) or DeepSeek-R1 \cite{Nature:DeepSeek} (\#8), the BERTScores are 0.9105 or 0.8980, demonstrating more conservative or more radical translation strategies, respectively. However, SR decreases by 5 or 4 percentage points and NE increases by 0.69 or 0.27 meters, respectively, which suggests that overly conservation may bring tedious noises while overly radical may delete essential semantics. 
Compared to them, GPT-4.1 \cite{gpt} (\#9) achieves the best trade-off. Therefore, we select GPT-4.1 as the LLM for the instruction translator module in T-araVLN.

\begin{table}[!t]
\caption{Ablation study results on the five translation principles in the instruction translator.}
\label{tab:ablation_experiment}
\vspace{2.0ex}

\centering
\resizebox{0.75\linewidth}{!}{
\renewcommand{\arraystretch}{1.1}
\begin{tabular}{rl ccc}
\toprule
\# & \textbf{Translation Principle} & \textbf{SR}$\uparrow$ & \textbf{NE}$\downarrow$ & \textbf{BERTScore} \\ 
\midrule
11 & Baseline                          & \underline{0.47} & \underline{2.91} & -  \\
12 & \quad + Inessential Removal       & 0.50 & 2.80 & 0.9041 \\
13 & \quad + Errors Revision           & 0.51 & 2.85 & 0.9028 \\
14 & \quad + High-Low Separation       & 0.50 & 2.90 & 0.9052 \\
15 & \quad + Representational Rotation & 0.49 & 2.80 & 0.9009 \\
16 & \quad + Representational Movement & 0.57 & 2.39 & 0.9005 \\
\rowcolor{gray!15} 17 & T-araVLN (Ours)& \textbf{0.63} & \textbf{2.28} & 0.9035 \\
\bottomrule
\end{tabular}
}

\footnotesize 
\vspace{2.0ex}
$\uparrow$ and $\downarrow$ indicate that higher and lower values correspond to better performance, respectively. 
“+” represents integration. 
\textbf{Bold} and \underline{underline} mark the best and worst scores, respectively.

\end{table}

\begin{figure}[!t]
  \centering
  \begin{subfigure}[b]{0.4\columnwidth}
    \includegraphics[width=\linewidth]{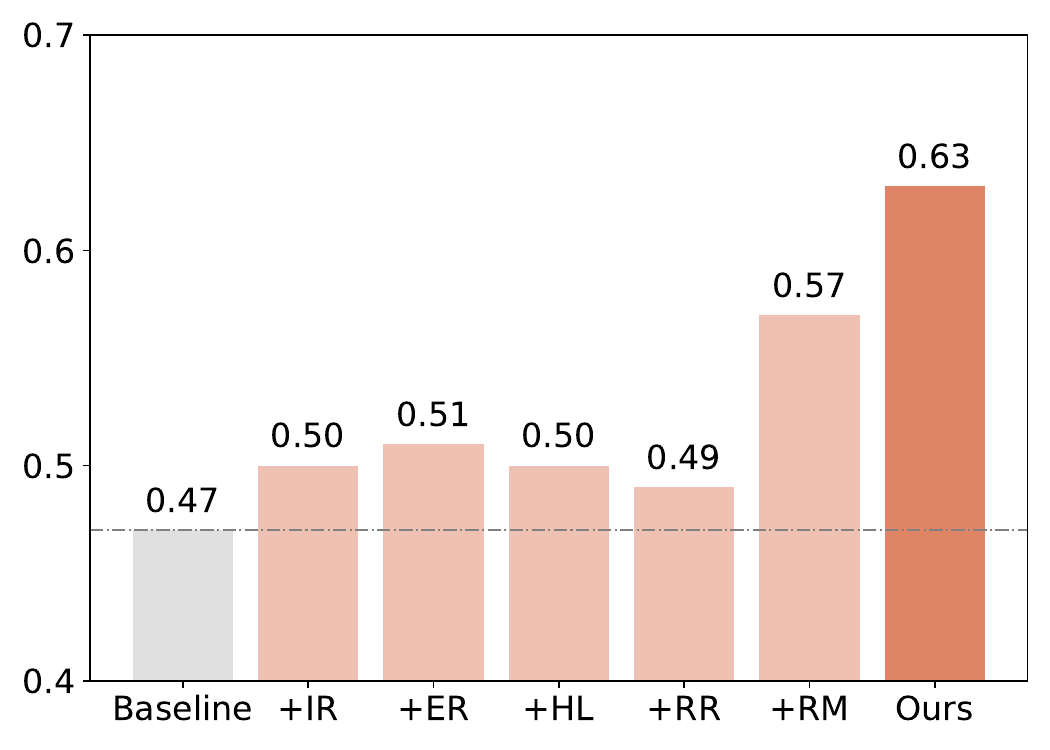}
    \caption{Success Rate $\uparrow$}
  \end{subfigure}
  \begin{subfigure}[b]{0.4\columnwidth}
    \includegraphics[width=\linewidth]{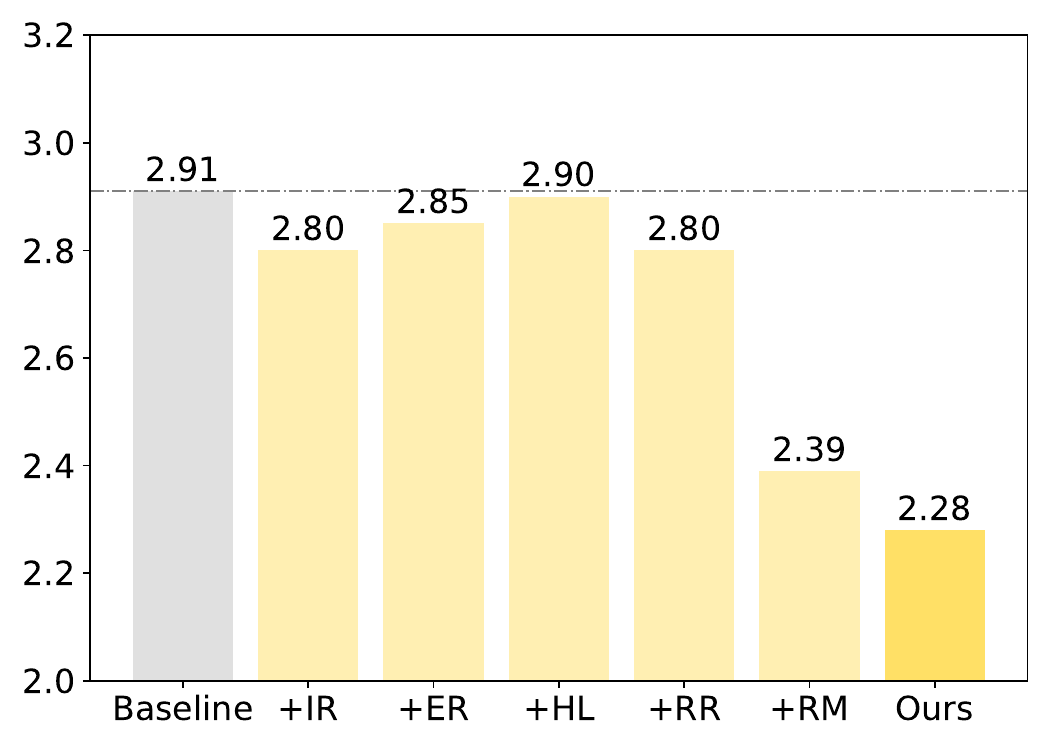}
    \caption{Navigation Error $\downarrow$}
  \end{subfigure}
  \caption{
  Ablation study illustrations on the five translation principles 
  in the instruction translator.
  $\uparrow$ and $\downarrow$ indicate that higher and lower values correspond to better performance, respectively. 
  “+” represents integration. 
  “$\text{-\,-\,-}$” marks the baseline score.
  }
  \label{fig:ablation_chart}
\end{figure}

\begin{figure*}[t]
\centering
\includegraphics[width=1.0\linewidth]{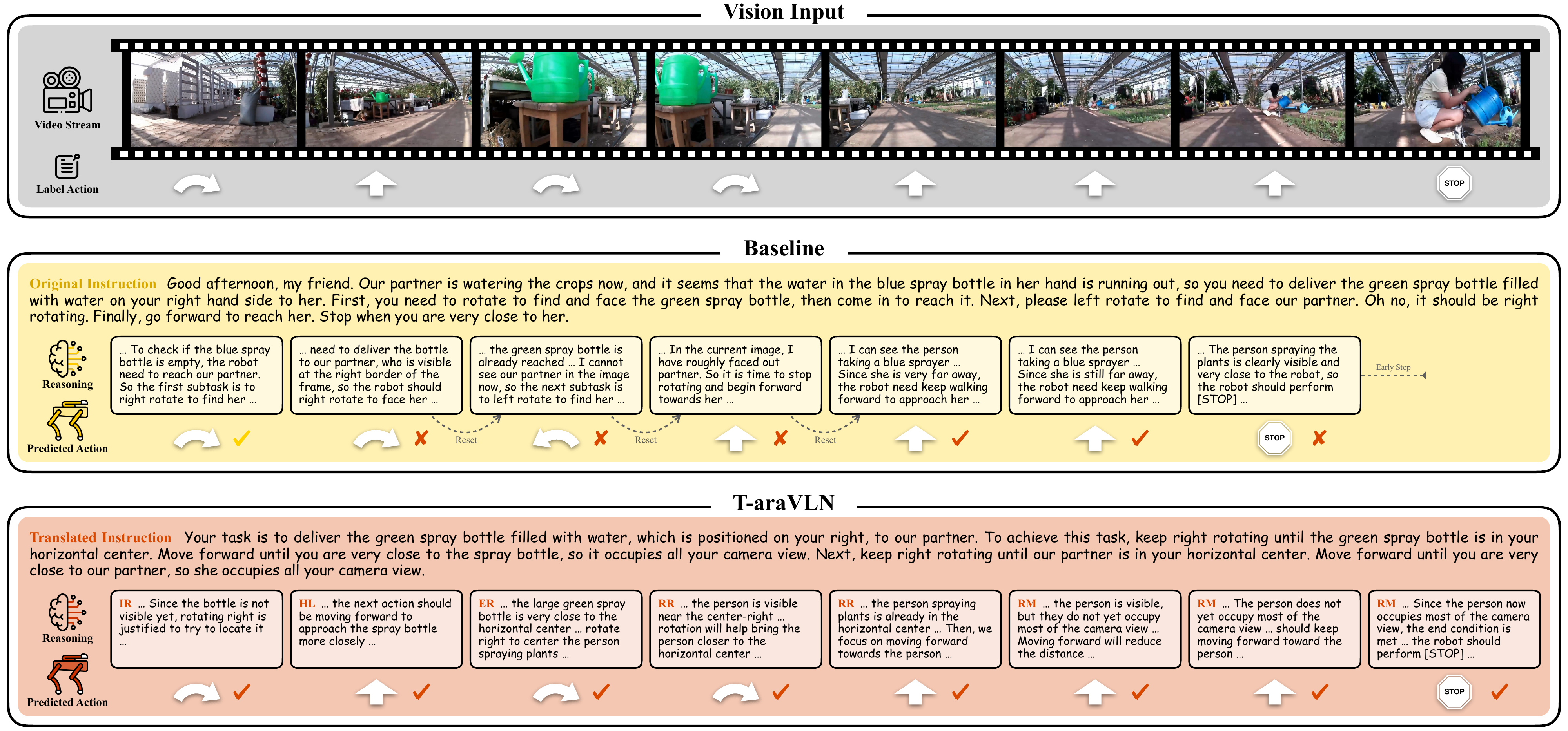}
\caption{
  Qualitative experiment illustration on a representative episode: 
  The original instruction includes several noises and mistakes, resulting in the challenging alignment between linguistic and visual inputs. Our T-araVLN, however, translates the instruction to be refined and precise, effectively improving the alignment performance. \textcolor[HTML]{CC3300}{\checkmark} represents correct prediction after rational reasoning; \textcolor[HTML]{FFCC00}{\checkmark} represents correct prediction after illogical reasoning; \textcolor[HTML]{CC3300}{\ding{55}} represents wrong prediction after illogical reasoning. Only for a complete demonstration, the baseline model is reset to the ground-truth path after deviation. (Zoom in for a better observation)}
\label{fig:qualitative_experiment}
\end{figure*}

\subsubsection{Translation Principle}
We ablate the five translation principles in the instruction translator module, in which we follow GPT-4.1 as the LLM, as shown in Table \ref{tab:ablation_experiment} and Figure \ref{fig:ablation_chart}. When every single principle is applied (\#12, \#13, \#14, \#15 or \#16), the performance improves by 2 $\sim$ 10 percentage points on SR and 0.01 $\sim$ 0.52 meters on NE, which indicates the effectiveness of every principle. When all five principles are combined (\#17), the performance reaches the best, i.e., our proposed T-araVLN, which indicates that every principle contributes to a different emphasis of translation, meanwhile, all five principles collaborate without significant interferences.


\subsection{Qualitative Experiment}

We implement the case study on a representative episode, as illustrated in Figure \ref{fig:qualitative_experiment}, in which we select a total of eight pivotal time steps for demonstration.


On the third time step, the relative instruction part is \textit{“reach it (the green spray bottle). Next, please \underline{left rotate} to find and face our partner. Oh no, \uwave{it should be right rotate}”}. From the camera image, we can observe that the green spray bottle is already reached, so the ground-truth action is starting \texttt{[RIGHT ROTATE]} to find and face the partner. 

Unfortunately, there is a speaking error in the instruction. 
The baseline only recognize the action content mentioned first (marked by \underline{underline}), but neglect the revision content mentioned later (marked by \uwave{uwave}). 
Consequently, it mistakenly thinks that \textit{the next subtask is to left rotate to find her}, resulting in the wrong \texttt{[LEFT ROTATE]} prediction.


In our T-araVLN, however, the instruction translator module successfully recognizes this speaking error, and reasonably revises the corresponding instruction part to \textit{“keep right rotating until our partner is in your horizontal center”} according to the ER principle. This translation effectively helps the decision-maker properly chooses to \textit{rotate right to center the person}, leading to the proper \texttt{[RIGHT ROTATE]} prediction.

From this case and the above experiments, we suggest an answer to the question mentioned at the beginning: Yes, we can. Our proposed instruction translator module can effectively polish an instruction to make it easier for the decision-maker to understand, supporting the state-of-the-art performance of T-araVLN.



\section{Conclusion}
\label{sec:conclusion}
\par In this paper, we present the T-araVLN method, in which we propose the instruction translator module to translate an instruction to be more refined and precise. When evaluated on the A2A \cite{arXiv:AgriVLN} benchmark, our T-araVLN effectively improves SR from 0.47 to 0.63 and reduces NE from 2.91m to 2.28m, demonstrating the state-of-the-art performance in the agricultural domain. 
\par During the experiments, we also find three main limitations: 
1) The visual features are not utilized to assist the instruction translation; 
2) The reasoning time cost of the translator on a single instruction is about 0.8s $\sim$ 2.5s; 
3) All the LLMs and VLMs are accessed only through APIs.
Hence, we suggest further improvements on multi-modal fusion, 
lightweight computational cost, 
and open-source model transferring.
\par In the future, in addition to the above improvements on existing limitations, 
we plan to further deploy T-araVLN on the Unitree Go2 Air \cite{Unitree} four-legged robotic dog, 
to extend the performance evaluation from simulator to reality.

\section*{Acknowledgment}
This study is supported by the National Natural Science Foundation of China's Research on Distributed Real-Time Complex Event Processing for Intelligent Greenhouse Internet of Things (Grant No. 61601471). 
Thanks to Thailand, Myanmar, and Sri Lanka for the impressive traveling experiences, giving us a chilled vibe for experiments and writing. 
Thanks to Yuanquan Xu, the inspiration to us.

\bibliographystyle{splncs04}
\bibliography{refs}

\end{document}